\documentclass[letterpaper]{article} 
\usepackage{aaai23}
\usepackage{times}  
\usepackage{helvet}  
\usepackage{courier}  
\usepackage[hyphens]{url}  
\usepackage{graphicx} 
\usepackage{natbib}  
\usepackage{caption} 
\usepackage{algorithm}
\usepackage{algorithmic}
\usepackage{amsmath}
\usepackage{newfloat}
\usepackage{listings}
\DeclareCaptionStyle{ruled}{labelfont=normalfont,labelsep=colon,strut=off} 
\floatstyle{ruled}
\newfloat{listing}{tb}{lst}{}
\floatname{listing}{Listing}

\usepackage{mathtools}
\usepackage{amsthm}
\newtheorem{example}{Example}
\newcommand\inter[2]{\mathcal{I}_{\mathcal{#1}}(#2)}

\usepackage{subcaption}
\title{PaTeCon: A Pattern-Based Temporal Constraint Mining Method for Conflict Detection on Knowledge Graphs}
\author{
    \\
    Jianhao Chen\equalcontrib, 
    Junyang Ren, 
    Wentao Ding\equalcontrib, 
    Yuzhong Qu 
}
\affiliations{
    
    State Key Laboratory for Novel Software Technology, Nanjing University, China\\
    \{jh\_chen, jyren, wtding\}@smail.nju.edu.cn, yzqu@nju.edu.cn
}

\begin{document}
\maketitle
\begin{abstract}

 Temporal facts, the facts for characterizing events that hold in specific time periods, are attracting rising attention in the knowledge graph (KG) research communities. In terms of quality management, the introduction of time restrictions brings new challenges to maintaining the temporal consistency of KGs and detecting potential temporal conflicts. Previous studies rely on manually enumerated temporal constraints to detect conflicts, which are labor-intensive and may have granularity issues. We start from the common pattern of temporal facts and constraints and propose a pattern-based temporal constraint mining method, PaTeCon. PaTeCon uses automatically determined graph patterns and their relevant statistical information over the given KG instead of human experts to generate time constraints. Specifically, PaTeCon dynamically attaches class restriction to candidate constraints according to their measuring scores.We evaluate PaTeCon on two large-scale datasets based on Wikidata and Freebase respectively. The experimental results show that pattern-based automatic constraint mining is powerful in generating valuable temporal constraints.

\end{abstract}

\section*{Introduction}
\label{sec:intro}


Knowledge graphs (KGs) represent real-world facts and support countless AI applications in many areas, such as information retrieval, natural language question answering, and healthcare. 
Classical KGs provide a static view of the real world by RDF triples (i.e., subject-predicate-object)
, which cannot satisfy the downstream application's demands of the specific occurring time periods of events.
For example, classical KGs may simply record Backham's career in Real Madrid CF as (\textsf{David\_Beckham, play\_for, Real\_Madrid\_CF}). Such facts are not sufficient enough to answer questions like ``When did David Beckham play for Real Madrid CF?''.
The demand for modeling and acquiring temporal facts for KGs has attracted rising attention in recent years. 
Formally, a temporal fact can be represented as a quadruple $(S,P,O,T)$, where time interval $T$ represents the time period in which the subject-predicate-object triple holds.

In practice, some popular KGs such as Freebase and Wikidata have introduced many temporal facts with particularly designed representation structures. However, the practical KGs are far from complete, and have quality issues.  Example~\ref{eg:conflict} illustrates a real conflict in Wikidata.
\begin{example}\label{eg:conflict}
Wikidata, one of the most widely used KG, contains two temporal facts as follows:
\begin{align*}
&\left(\mathsf{Q718275},\mathsf{member\_of\dots},\mathsf{Q180798},[1998,2009]\right),\\
&\left(\mathsf{Q718275},\mathsf{member\_of\dots},\mathsf{Q180798},[2002,-]\right),
\end{align*}, where $\mathsf{Q718275}$ and $\mathsf{Q180798}$ are Wikidata ids of ``Juli Fernández'' and ``Andorra national football team'' respectively.
It is obvious they can't both hold because ``\textit{two different career experience of an athlete should be disjoint in time}''.
\end{example}
The current state of practical KGs brings the need of acquisition and quality management of temporal knowledge. 
The introduction of time restrictions brings a new challenge to KG quality management~--~the inconsistency in the time dimension. 
Besides, deep learning-based knowledge acquisition has become a very popular trend in recent years. Since they cannot provide theoretical guarantees of the consistency of the acquired knowledge, explicit quality checks on the acquired knowledge may be needed.

In order to maintain the temporal consistency of KGs, we need to find the conflicting temporal facts. As shown in the previous example, a common perception in temporal conflict detection is that violations of specific constraints cause people to believe that certain temporal facts are conflicting. Example~\ref{eg:raw_rule_sports_teams} illustrates such a rule found in \citet{AAAI17}'s work.
By enumerating the temporal constraints on a given knowledge graph, we can easily identify all potential conflicts that violate these constraints.

\begin{example}
\label{eg:raw_rule_sports_teams}
Rule for representing the constraint ``\textit{One can't be a member of two sports teams at the same time}'':
\begin{align}
\mathsf{disjoint}\left(t_1,t_2\right) \coloneq  &
\left(x,\mathsf{member\_of\_sports\_team},y,t_1\right), \notag\\
& \left(x,\mathsf{member\_of\_sports\_team},z,t_2\right), \notag\\
& y\neq z  \label{f:eq1}.
\end{align}
\end{example}

However, previous studies enumerate the temporal constraints by human experts. Although experienced experts can provide high-quality constraints, manual enumeration of temporal constraints is labor-intensive, and human experts may have difficulty on judging the granularity of temporal constraints.
For example, \citet{AAAI17}'s constraint in formula~\ref{f:eq1} fails to note that the range of the $\mathsf{member\_of\_sports\_team}$ property is actually $\mathsf{sports\_organization}$, which encompasses both national teams and clubs, and that it would mistake ``an athlete who is a member of a club while being a member of the national team'' as a temporal conflict. Manually distinguishing these exceptions will be tedious, if not infeasible.
Moreover, manual constraint engineering lacks generality. The pre-determined constraints can neither be automatically adopted when the KG is updated nor directly migrated to other KGs. Therefore, there is a need for an efficient and general automatic constraint mining technique.

\begin{figure}[!h]
    \centering
    \subcaptionbox{}{\includegraphics[scale=0.7]{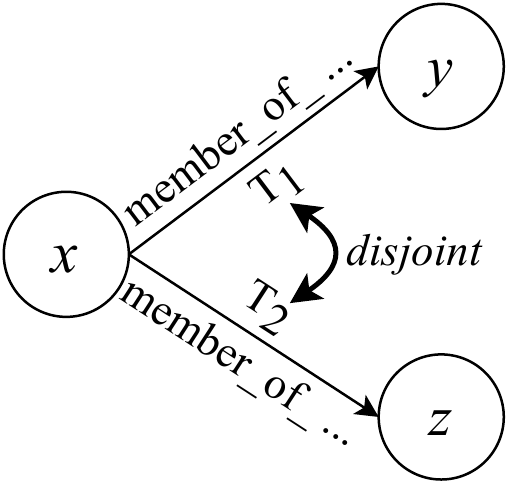}}
    \subcaptionbox{}{\includegraphics[scale=0.7]{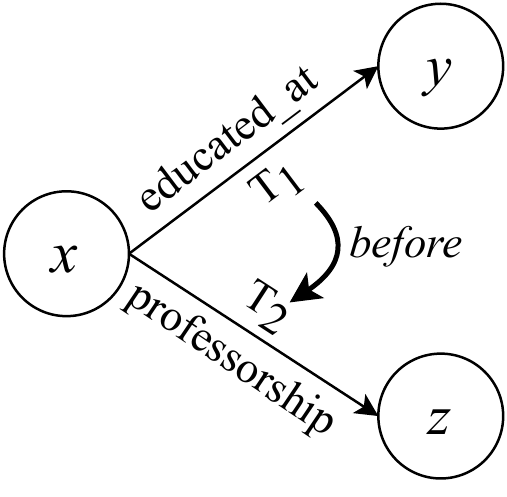}}
    \caption{Examples of the graph representation of temporal constraints. Note that different variables (e.g., $y$ and $z$) should correspond to different things if not specified.}
    \label{fig:2tc}
\end{figure}
\begin{figure}[!h]
    \centering
    \includegraphics[scale=0.7]{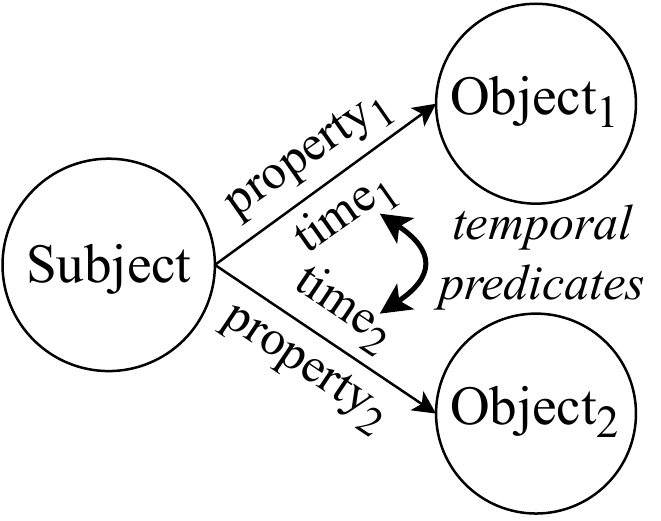}
    \caption{The common structural pattern of the temporal constraints illustrated in Figure~\ref{fig:2tc}.}
    \label{fig:cSP}
\end{figure}

As illustrated in formula~\ref{f:eq1}, the existing studies suggest that temporal constraints can be modeled as logical rules. Our observation further shows that the rule can be divided into graph patterns of KG facts and temporal assertions over the facts. For example, the body part of the rule in formula~\ref{f:eq1} matches two facts with the same subject, predicate, but different objects. Its head part is a temporal assertion of the disjointness of the involved time intervals. Detection of temporal conflicts is the process of finding all subgraphs that can match the patterns. 
This perspective draws our attention to common patterns in constraints themselves. For example, the temporal constraint on athletes and another constraint ``one must be educated somewhere before his professorship'' can be modeled as illustrated in Figure~\ref{fig:2tc} respectively, and they share a common structural pattern, as illustrated in Figure~\ref{fig:cSP}.

Therefore, we propose a pattern-based method for mining temporal constraints, \textbf{PaTeCon}\footnote{The released source code and documentation are available at \url{https://github.com/JianhaoChen-nju/PaTeCon}}. We summarize the existing work and propose two structural patterns for temporal constraints, which are presented in the preliminaries section~\ref{sec:sp} in detail. We utilize these structural patterns to get the actual graph pattern in the given KG. For each graph pattern, we attach temporal predicates to the corresponding time intervals to generate candidate constraints, measure the quality of the obtained candidates, and use the high-quality constraints to detect temporal conflicts. In particular, we dynamically attach class restrictions to temporal constraints according to their measuring scores to better determine the granularity of the temporal constraints.


The remainder of the paper is organized as follows. The second section introduces the preliminaries, including the definition of temporal facts, constraints, and their structural patterns. The third section presents the framework and our implementation of PaTeCon in detail. The fourth section illustrates the evaluation of PaTeCon with two novel large-scale datasets. The fifth section describes the related work. The last section summarizes our main contributions and concludes this paper.

\section*{Preliminaries}
\subsection*{Knowledge Graph with Temporal Facts}
In this paper, the term \textit{knowledge graph (KG)} refers to knowledge bases represented with the RDF data model. A classical RDF KG expresses knowledge via a set $\mathcal{R}$ of IRI resources (resources with identifiers) and a set $\mathcal{L}$ of literals (string data). The conceptual terms of the KG are modeled by classes $\mathcal{C} \subseteq \mathcal{I}$ and properties $\mathcal{P} \subseteq \mathcal{I}$ of the IRI resources.
An atomic fact in the RDF KG is a triplet $F=(S, P, O) \in \mathcal{R}\times\mathcal{P}\times (\mathcal{R} \cup \mathcal{L})$. RDF also allows representing complex facts (e.g., facts with time restrictions) via anonymous resources, aka. RDF blank nodes. 
To express the theoretical model of the time constraints mining, we use the term \textit{temporal facts} to denote the complex facts with time restrictions. A temporal fact is a quadruple $TF=(S, P, O, T) \in \mathcal{R}\times\mathcal{P}\times (\mathcal{R} \cup \mathcal{L})\times (\mathcal{T} \times \mathcal{T})$, where $\mathcal{T}$ denotes the time domain and $T=(t.s, t.e)$ represents a time interval defined by its start and end times.

\subsection*{Time Intervals and Their Algebra}
A time interval $T=(t.s, t.e)$ is represented as an ordered pair of time values $t.s$ and $t.e$, denoting its \textit{start} and \textit{end} points respectively.  By comparing the endpoints of two time intervals, we can summarize their relation to interval algebras such as \citet{ALLEN83}. According to the practical demands of generating temporal constraints, we use 5 temporal predicates including $\mathsf{start}$, $\mathsf{finish}$, $\mathsf{before}$, $\mathsf{disjoint}$, and $\mathsf{include}$. The calculation details of these predicates are presented in the method section~\ref{sec:ut}.

\begin{figure*}[t]
    \centering
	\subcaptionbox{}{\raisebox{0.075\height}{\includegraphics[scale=0.75]{figures/pattern_structure_2.pdf}}}
	\hspace{30mm} 
	\subcaptionbox{}{\includegraphics[scale=0.75]{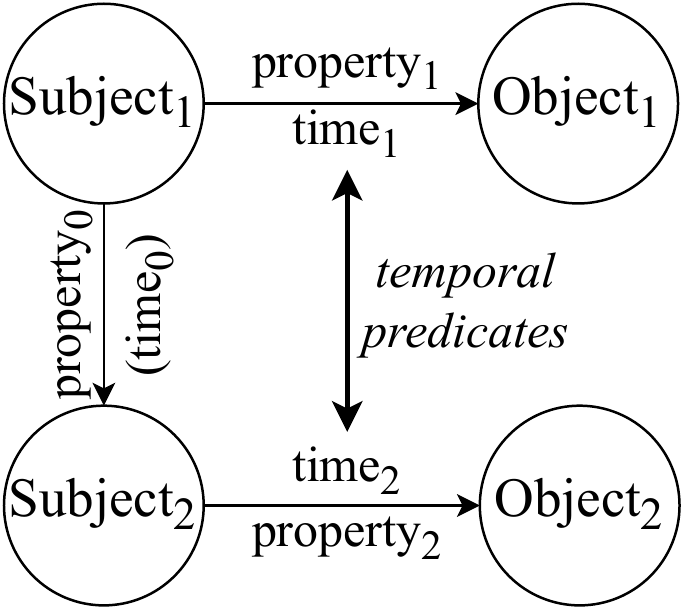}}
    \caption{Structural patterns. Noting that the properties could be reversed.}
    \label{fig:structural patterns}
\end{figure*}

\subsection*{Structural Patterns}\label{sec:sp}
In this paper, we model temporal constraints via their common patterns, i.e., the structural patterns. Structural patterns portray the graph structure of facts in certain constraints and indicate which resources in the facts are involved in the temporal predicates. PaTeCon uses two basic structural patterns, as illustrated in Figure~\ref{fig:structural patterns}. 
These structural patterns can capture the temporal relationships between two subjects or two experiences of one subject. As previously illustrated in the introduction, structural patterns can model various types of constraints by changing the filling-in predicates.

Specifically, structural pattern (a) can model most of the temporal relations that previously studies concern, such as temporal disjointness, precedence and inclusion \cite{AKBC14,AAAI17,ETCwise18}. It can also model temporal constraints that are not interval relations. For example, a common type of constraint in previous studies is \textit{mutual exclusion}, which expresses the uniqueness of certain events. Example~\ref{eg:me} illustrates how to express such constraints by structural pattern (a).
\begin{example}\label{eg:me}
``One can only be born once'', thus one person should have only one birthplace. This constraint can be modeled via structural pattern (a) by taking $\mathsf{false}$ as a special atom in the predicate slot.
\begin{align}
\mathsf{false} \coloneq & \left(x,\mathsf{place\_of\_birth},y,t_1\right),\notag\\
&\left(x,\mathsf{place\_of\_birth},z,t_2\right),\notag\\
&y\neq z \label{f:eq3}.
\end{align}
\end{example}

\begin{example}\label{eg:patternB}
``A Ph.D. graduates apparently after his doctoral advisor's educational experience''. This constraint can be modeled via structural pattern (b) by filling $\mathsf{doctoral\_advisor}$, $\mathsf{academic\_degree}$ and $\mathsf{educate\_at}$ into property 0,1,2. 
\begin{align}
\mathsf{before}\left(t_1,t_2\right) \coloneq & \left(x,\mathsf{doctoral\_advisor},y\right),\notag\\
&\left(y,\mathsf{educate\_at},z,t_1\right),\notag\\
&\left(x,\mathsf{academic\_degree},doctorate,t_2\right)
\label{f:patternB}
\end{align}
\end{example}

Structural pattern (b) further considers the assertions about multiple subjects. It assumes that two temporally-related subjects must be connected in the KG. Example~\ref{eg:patternB} illustrates how to express such constraints by structural pattern
(b). To limit the search space to a controllable size, we do not mine the disjointness relations between two different subjects in the practical implementation. More complex constraints, such as constraints over three subjects, can be generated by combining the constraints of simple patterns. So we do not create structural patterns for them.

\subsection*{Temporal Constraints}
A temporal constraint is a logical assertion of the time dimension, such as illustrated in formula~\ref{f:eq1}. 

The structural pattern perspective shows that a temporal constraint can be divided into two parts, 1) the triple/quadruple patterns for facts in KGs and 2) the constraint predicate over the facts. We formalize a constraint $tc$ in the form $tc = H\coloneq B$, where head $H$ is an atomic predicate, and the body $B$ is a conjunction of atoms with optional variables. 
In practical applications, we use the quadruples to match with temporal facts in KGs. For example, the quadruple 
\begin{align}
\left(\mathsf{Juli\_Fern\acute{a}ndez},\mathsf{member\_of\_sports\_team},x,T\right)
\end{align}
can matches with all of Juli Fernández's experiences as an athlete. We use $\mathcal{I}_{\mathcal{G}}(tc)$ to denote the subgraphs of $\mathcal{G}$ satisfying the body (i.e., the graph pattern) of $tc$.

\section*{Method}
Algorithm~\ref{alg:algorithm} describes the framework of our method, PaTeCon. The method first mines frequent candidate constraints according to the pre-defined structural patterns $\mathcal{SP}$, then computes the $confidence$ scores of the candidates to dynamically refine them and determines the final set of mined temporal constraints. 

The next following subsections describe the key details in candidates mining, confidence computation and constraint refinements.

\begin{algorithm}[ht]
\caption{Temporal Constraint Mining}
\label{alg:algorithm}
\textbf{Input}:the KG $\mathcal{G}$, structural patterns $\mathcal{SP}$
and temporal predicates $\mathcal{TP}$\\
\textbf{Output}:the constraints $\mathcal{TC}$
\begin{algorithmic}[1]
\STATE $Candidates = \emptyset$ \hfill\textit{// Mining candidate constraints.}
\STATE $\mathcal{GP} = \mathrm{InstantiateToGraphPatterns}(\mathcal{G},\mathcal{SP})$
\FOR{$gp \in \mathcal{GP}$}
    \STATE $sg = \mathrm{MatchedSubGraphs}(\mathcal{G},gp)$
    \FOR{$tp \in \mathcal{TP}$}
        \STATE $tc = \mathrm{MakeConstraint}(tp, gp)$
        \STATE $s = \mathrm{ComputeSupport}(sg, tc)$
        \IF{$s < \theta_{freq}$}
            \STATE \textbf{continue}
        \ENDIF
        \STATE $Candidates = Candidates \cup \{tc\}$
    \ENDFOR
\ENDFOR
\STATE $\mathcal{TC}=\emptyset$ \hfill\textit{// Obtaining high-quality constraints.}
\FOR{$tc \in Candidates$}
    \STATE $gp = tc.\mathrm{graph\_pattern}$ 
    \STATE $sg = \mathrm{MatchedSubGraphs}(\mathcal{G},gp)$
    \STATE $c = \mathrm{ComputeConfidence}(sg, tc)$
    \IF{$c > \theta_{c1}$}
        \STATE $\mathcal{TC}=\mathcal{TC} \cup \{tc\}$
    \ELSIF{$c > \theta_{c2}$}
        \STATE $\mathcal{RC} = \mathrm{RefineConstraint}(tc)$
        \FOR{$rc \in \mathcal{RC}$}
            \IF{$\mathrm{ComputeSupport}(sg, rc) > \theta_{freq} \wedge \mathrm{ComputeConfidence}(sg, rc) > \theta_{c1}$}
                \STATE $\mathcal{TC}=\mathcal{TC} \cup \{tc\}$
            \ENDIF
        \ENDFOR
    \ENDIF
\ENDFOR
\STATE \textbf{return} $\mathcal{TC}$
\end{algorithmic}
\end{algorithm}

\subsection*{Mining Candidate Constraints}
Lines $1$ to $10$ describe the mining of frequent candidates. PaTeCon first instantiates the pre-defined structural patterns $\mathcal{SP}$ to graph patterns $\mathcal{GP}$ that have actually appeared on the KG, i.e., fills the \textit{property} slots in $SP$s according to the connection structure of facts in $\mathcal{G}$. For the generated graph patterns, PaTeCon attempts to attach temporal predicates to them to build temporal constraints. The constraints whose support $s$ is no less than a threshold $\theta_{freq}$ will be considered as candidates. Specifically, the instantiation is accomplished by taking all entities in $\mathcal{G}$ as possible subjects in $SP$s and searching their neighboring subgraphs to obtain the properties.



\subsection*{Measuring Constraints}\label{sc:measure}
The quality of a temporal constraint $tc$ on the given KG $\mathcal{G}$ can be measured on $\inter{G}{tc}$, the facts that match with the corresponding graph pattern. 
We measure the quality of a constraint by its \textit{entity-level confidence}, which is defined as follows:
\begin{align}
confidence = \frac{\#entities_{pos}}{\#entities_{pos} + \#entities_{neg}},\label{f:conf}
\end{align}
where $entities_{pos.}$ and $entities_{neg.}$ are the subsets of subject entities in $\inter{G}{tc}$. Specifically, we have
\begin{align}
entities = \bigcup_{F \in \inter{G}{tc}} subject(F),
\end{align}
where $F$ denotes the subgraphs matched by $tc$'s body, $subject(F)$ denotes the entities in the $F$ that matched by the $subject$ and $subject_1$ slots in structural pattern (a) and (b) respectively.

In the classification of entities, an important issue is the uncertainty of KGs in the time dimension. Specifically,
practical KGs cannot provide arbitrarily accurate occurring periods of events. For example, Wikidata records that both the $\mathsf{start\_time}$ and $\mathsf{end\_time}$ of Barack Obama's employment as a Sidley Austin employee are 1991, which doesn't mean that the $\mathsf{start\_time}$ and $\mathsf{end\_time}$ are precisely equal, nor does it mean that Obama held the job throughout the whole 1991. Besides, because of the incompleteness of practical KGs, some facts that indicate ended events have no $\mathsf{start\_time}$ or $\mathsf{end\_time}$.
Therefore, we classify $entities$ to the positives $entities_{pos}$, the negatives $entities_{neg}$, and the unknowns $entities_{unk}$ according to the logical value of the head predicate in $tc$. An entity is considered positive if all the logical values of the matched subgraphs are positive and considered unknown if any of the logical values of the matched subgraphs is unknown. We use the positives and the negatives to compute the confidence scores, and the unknowns are dropped.

The details about computing the logical values of temporal predicates are presented as follows:

\subsubsection*{Temporal Predicates over Uncertain Time Intervals}\label{sec:ut}
The logical values of the head predicates are computed over time intervals of the form $T=(t.s, t.e)$, where $t.s$ denotes the start time and  $t.e$ the end time. The positive conditions of the predicates are illustrated in Table~\ref{tab:ctp}. For cases where the granularity is not fine enough or where some of the time values are absent, we will set the computation result to \textit{unknown} instead of \textit{negative}, as illustrated in Table~\ref{tab:comp}.

\begin{table}[t]
    \centering
\begin{tabular}{l|l}
    \hline\hline
    Predicates & Positive Condition\\
    \hline
    $\mathsf{start}$ & $T_1.s = T_2.s$\\
    $\mathsf{finish}$ & $T_1.e = T_2.e$\\
    $\mathsf{before}$ & $(T_1.e < T_2.s) \vee (T_1.e = T_2.s \wedge T_1 \ne T_2)$\\
    $\mathsf{disjoint}$ & $(T_1 \,\mathsf{before}\, T_2) \vee (T_2 \,\mathsf{before}\, T_1)$\\
    $\mathsf{include}$ & $(T_1.s \le T_2.s) \wedge (T_2.e \le T_1.e)$\\
    \hline\hline
\end{tabular}
    \caption{Calculation of temporal predicates.}
    \label{tab:ctp}
\end{table}

\begin{table}[b]
    \centering
    \begin{tabular}{c|c|c|c}
        \hline\hline
        $t_1$ & $t_2$ & $t_1 < t_2$ & $t_1 = t_2$  \\
        \hline
        2021-12 & 2022 & Positive & Negative \\ 
        2022-01 & 2022 & Unknown & Unknown \\ 
        - & 2022 & Unknown & Unknown \\
        \hline\hline
    \end{tabular}
    \caption{Calculation on (possibly absent) time values of different granularity.}
    \label{tab:comp}
\end{table}

\subsection*{Refining Constraints}
As previously mentioned in the introduction section, constraints with only property restrictions may be too coarse for conflict detection. If the confidence $c$ of a constraint $tc$ is lower than the quality threshold $\theta_{c_1}$ but is beyond a more relaxed threshold $\theta_{c_2}$, we will attempt to restrict the domain of it by class in $\mathcal{G}$. Specifically, we will attempt to enumerate the combination of classes of corresponding entities in the matched subgraphs $\mathcal{SG}$. All refined constraints whose support is no less than $\theta_{freq}$ and confidence is no less than $\theta_{c_1}$ will be added to the final results.

\section*{Evaluation}
\subsection*{Experimental Setup}

\subsubsection{Datasets}
We use \citet{AAAI17}'s 50k samples from Wikidata (denoted as WD50K) to comparing PaTeCon with it. To bring our evaluations closer to practical scenarios while ensuring a manageable scale, we also constructed two datasets, WD27M and FB37M, with two different KGs respectively.
For WD27M, we extend the property list to obtain more temporal properties. We collect all the properties in facts with time qualifiers (including $\mathsf{point\_in\_time}$, $\mathsf{start\_time}$, and $\mathsf{end\_time}$) to obtain a set of 93 temporal properties and extract 141 properties about person and organizations as non-temporal property list. We extract all facts with the 93+141 properties from the 2019-01-28 dumps of Wikidata\footnote{\url{https://archive.org/download/wikibase-wikidatawiki-20190128}} for evaluation. 
Since Freebase's schema does not give an explicit list of time qualifiers, we follow \citet{AAAI17} to use 6 temporal properties for FB37M, we collect the entities that these properties described, collecting all their one-hop facts in the latest dumps of Freebase\footnote{\url{https://developers.google.com/freebase}} for evaluation. The statistics about the three datasets are illustrated in Table~\ref{tab:datasets}.

\begin{table}[t]
    \centering
    \begin{tabular}{l|r|r|r}
        \hline\hline
        Datasets & WD50K & WD27M & FB37M\\
        \hline
        Entities & 17,176 & 7,224,869 & 10,178,664\\
        Facts & 50,000 & 27,312,354 & 37,939,422\\
        \quad Temporal & 50,000 & 7,471,929 & 2,989,799 \\
        \quad Others & 0 & 19,840,425 & 34,949,623\\
        Properties & 6 & 234 &  2,490\\
        \quad Temporal & 6 & 93 & 6 \\
        \quad Others & 0 & 141 & 2,484\\
        \hline\hline
    \end{tabular}
    \caption{Dataset Overview}
    \label{tab:datasets}
\end{table}

\subsubsection{Evaluation Metrics and Setting}
We classify the quality of temporal constraints into 3 grades as follows: 
\begin{itemize}
\item \textbf{C}: The constraint is \underline{c}orrect.
\item \textbf{M}: The constraint has some \underline{m}erit, but there are also obvious exceptions.
\item \textbf{W}: The constraint is \underline{w}rong. The involved facts are not temporally related or should be restricted by an opposite predicate.
\end{itemize}
In our evaluation, each constraint is scored by 3 experienced annotators separately. 
We will simply set the quality to \textbf{M} in the only case where consensus could not be reached -- where 3 annotators rate the constraint as \textbf{C}, \textbf{M}, and \textbf{W} respectively. 
Specifically, we evaluate all the generated constraints on the dedicated dataset, WD50K. For the practical large-scale datasets, we randomly sample 50 constraints for evaluation. We report the quality rates and the number of detected possible conflicts of PaTeCon. We also sample 50 possible conflicts to investigate whether the detected subgraphs are truly conflicting.

\subsubsection{Environments and Parameters}
All the results of PaTeCon are obtained by a single-threaded Python implementation on a personal workstation with an Intel Xeon CPU E5-1607 v4 @3.10GHz CPU and 128GB RAM. For the threshold values, we simply set $\theta_{freq}=20$ in WD50K and $\theta_{freq}=100$ in WD27M and FB37M, $\theta_{c_1}=0.5$ and $\theta_{c_2}=0.9$ in all datasets. 


\subsection*{Main Result}

\subsubsection{Mined Constraints vs. Hand-crafted Constraints}
\begin{table}[t]
    \centering
    \begin{tabular}{c|r|r|r|r}
    \hline\hline
        Method & C & M & W & Total\\ \hline
        \citet{AAAI17} & 9 & 3 & 0 &12\\
        PaTeCon & 13 & 0 & 0 & 13\\ \hline\hline
    \end{tabular}
    \caption{Statistics about the quality grades on WD50K.}
    \label{tab:WD50K}
\end{table}

Table~\ref{tab:WD50K} illustrates the comparison results of the PaTeCon mined constraints to the hand-crafted constraints in \citet{AAAI17}. 25\% of the hand-crafted constraints have quality issues while all the mined constraints are considered correct. The results show that PaTeCon performs better in specifying the granularity of constraints than human experts.

On WD50k, our statistics show that \citet{AAAI17}'s constraints regard about $10\%$ of all the facts combinations as possible conflicts. Specifically, their constraints detect 107,104 possible conflicts, and $96.8\%$ ($103, 662$) of which are included by the noisy constraint illustrated in Example ~\ref{eg:raw_rule_sports_teams}. In contrast, our method detects 971 conflicts, 967 of which are in \citet{AAAI17}’s results.

\begin{table}[t]
    \centering
    \begin{tabular}{c|r|r}
    \hline\hline
    & \#Constraints & \#Conflicts \\\hline
    WD27M & 644 & 703,715 \\\hline
    FB37M & 81 & 23,467 \\
    \hline\hline
    \end{tabular}
    \caption{Statistics about mined constraints and they detected possible conflicts on WD27M and FB37M.}
    \label{tab:c&c}
\end{table}

\begin{table}[t]
    \centering
    \begin{tabular}{c|r|r|r|r}
    \hline\hline
    & \#Cons. & C & M & W \\\hline
    WD27M & 644 & 50\% & 36\% & 14\%\\
    \quad Pat. (a) & 189 & 69\% & 31\% & 0\%\\
    \quad Pat. (b) & 455 & 41\% & 38\% & 21\%\\
    \hline
    FB37M & 81 & 58\% & 38\% & 4\%\\ 
    \quad Pat. (a) & 17 & 64\% & 36\% & 0\%\\
    \quad Pat. (b) & 64 & 56\% & 38\% & 5\%\\
    \hline\hline
    \end{tabular}
    \caption{Statistics about quality grades of PaTeCon mined rules on WD27M and FB37M.}
    \label{tab:cq}
\end{table}

\subsubsection{Performances on Practical Datasets}
Table~\ref{tab:c&c} illustrates the number of PaTeCon mined constraints and the possible conflicts they detected on large-scale datasets WD27M and FB37M. 
Table~\ref{tab:cq} further illustrates quality grades of the mined constraints by structural patterns. In summary, 86\% and 96\% of the PaTeCon mined constraints are considered valuable on WD27M and FB37M respectively, about half of the rules (50\% and 58\% on WD27M and FB37M respectively) of the PaTeCon mined constraints are considered correct and do not need further modifications. According to statistics by structural patterns, we can find that all the mined constraints of structural pattern (a) have some merits. structural pattern (b) will mistakenly treat some occasionally temporal relations as constraints, especially on Wikidata.

Specifically, the Fless's $\kappa$ score on WD50K, WD27M and FB37M are $0.10$, $0.29$ and $0.22$ respectively.

\subsubsection{Running Efficiency}
The running efficiency of PaTeCon is determined by the number of entities, properties, classes and the density of the given KG. Since practical KGs are very sparse, the execution time of PaTeCon is much less than the time to enumerate all combinations of properties and classes. Specifically, the cost of running our method on each benchmark is listed in Table \ref{tab:running_time}.

\begin{table}[t]
    \centering
    \begin{tabular}{c|r|r|r}
    \hline\hline
         & WD50K & WD27M & FB37M \\ \hline
        Constraint Mining & 5 & 2,072 & 10,647\\
        Conflict Detection & 2 & 1,080 &  763\\ \hline\hline
    \end{tabular}
    \caption{The running time (seconds) of PaTeCon on the three datasets.}
    \label{tab:running_time}
\end{table}

\subsection*{Case Study}
In the introduction section, we have mentioned that human-engineered constraints may have granularity issues. Example~\ref{eg:raw_rule_sports_teams} illustrates such a constraint in \cite{AAAI17}. We investigate the output of PaTeCon on WD27M. Our observation shows that PaTeCon can capture the original temporal constraint and refine it to more reliable sub-constraints. Example~\ref{eg:case0} illustrates one of the refined temporal constraints.

\begin{example}
\label{eg:case0}
The graph representation of an refined version of ``\textit{One can't be a member of two sports teams at the same time}'', mined from WD27M. 
\begin{figure}[htb]
    \centering
    \includegraphics[scale=0.7]{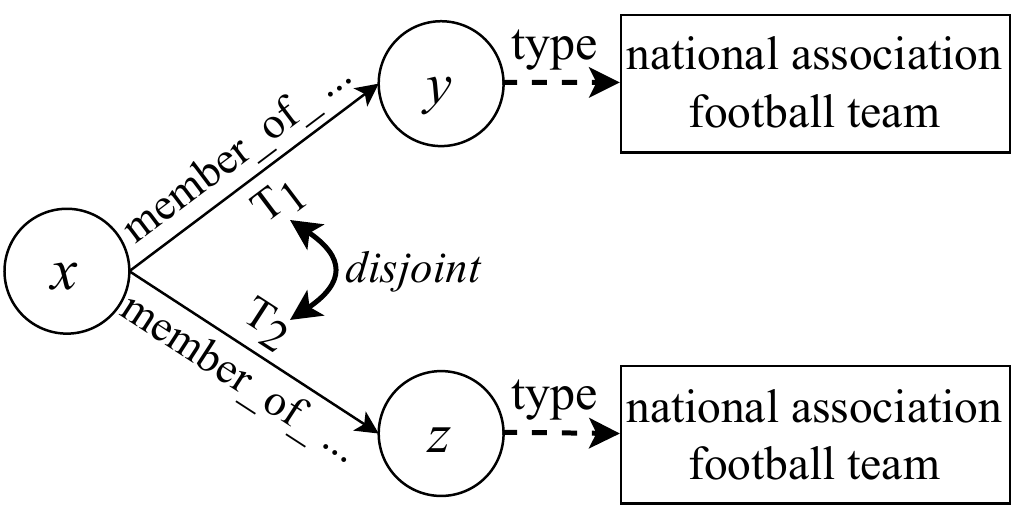}
    \caption{Graph representation of ``\textit{One can't be a member of two national association football teams at the same time}''.}
    \label{fig:case1}
\end{figure}
\end{example}

In summary, a total of 8 high-quality temporal constraints are generated, the class restriction over $y$ and $z$ including
$\mathsf{sports\_club}$, $\mathsf{national\_sports\_team}$, $\mathsf{UCI\_Continental\_Team}$, etc. This case shows that dividing coarse temporal constraints is necessary and can dramatically improve the quality of constraints.





\section*{Related Work}
\subsection*{Acquisition of Temporal Knowledge}
Acquisition of temporal knowledge is an active research topic in recent years. Commonly used temporal/event KGs include the Global Database of Event Language and Tone (GDELT), the Integrated Crisis Early Warning System (ICEWS), and Event-KG. The large-scale general KGs, such as Freebase and Wikidata, also include many temporal facts. Lots of researches focus on reasoning missing temporal facts via deep learning models \citep{Jiangcoling16,KNOWicml17,TTRANSEwww18,RENETemnlp20,LFHaaai21,CENacl22}. Some of the research benefited from the integration of the reasoning model with temporal constraints. \citet{Jiangcoling16} incorporate temporal consistency constraints for KG completion. \citet{TIMEPLEXemnlp20} extend their base model via additional (soft) temporal constraints containing relation recurrence, ordering between relations, and time gaps between relations. 

In summary, the widespread utilization of deep learning has achieved significant results and has placed efficiency demands on the design of explicit temporal constraints.

\subsection*{Temporal Conflict Detection and Resolution}
To the best of our knowledge, only a few work \cite{RTCbtw11,AKBC14,AAAI17,ETCwise18} has addressed temporal conflict detection and resolution. \citet{RTCbtw11} use first-order logic Horn formulas with temporal predicates to express temporal constraints. They employ a scheduling algorithm to resolve conflicts between facts. \citet{AKBC14} debug temporal knowledge graphs by computing the MAP state of a Markov Logic Network. \citet{AAAI17} present a Markov Logic Network (MLN) based approach for reasoning over temporal KGs with mostly hand-crafted temporal constraints and a few AMIE mined rules. AMIE, a general mining approaches for RDF KGs, is designed for mining rules about (S, P, O) triplets, thus it cannot handle the quadruple temporal facts where the time restriction is represented via an additional time interval. \citet{ETCwise18} manually define the constraint graph, which is actually a graph of temporal constraints, and then detect conflicting facts according to the constraint graph. They model the truth inference problem as a maximum weight clique problem. 

All of the above work highly relies on manual constraint engineering. Besides, their work assumes that the temporal facts sampled from practical KGs are correct. They construct dedicated datasets by adding incorrect facts and assigning confidence to each fact instead of detecting conflicts from practical KGs.

\section*{Conclusion}
This paper addresses the automatic mining of temporal constraints, where the temporal constraints are the logical assertions of subgraphs in KG on the time dimension. Specifically, we initially explore the idea of generating temporal constraints according to the statistics about the pre-defined patterns. We simply use 2 structural patterns with 5+1 predicates (5 temporal predicates and $\mathsf{false}$ for mutual exclusion) to demonstrate the effectiveness of the pattern-based constraint mining.
Our main contributions can be summarized as follows.
\begin{enumerate}
    \item We propose a pattern-based framework for automatic temporal constraint mining. Our framework does not rely on human efforts to obtain specific temporal constraints and can be generalized to arbitrary temporal KGs.
    \item We propose the constraint mining method \textbf{PaTeCon}. PaTeCon exploits statistical information of the instances of predefined structural patterns to mine candidate constraints.
    Specifically, it automatically refines coarse constraints by appending class restrictions.
    \item
    We propose two large-scale benchmarks, WD27M and FB37M, extending existing benchmarks in different dimensions for more practical evaluations. WD27M extends the list of temporal attributes and FB37M collects all the facts around the described entities.
\end{enumerate}

In real applications, we can extend PaTeCon to fit more complex demands. Specifically, the predicates can be extend to include quantitative relations such as $t_1 - t_2 \le P10Y$ ($t_1$ is at least 10 years before $t_2$). Extensions of PaTeCon, including the extension of predicates and the combination of mined constraints, are considered as future work. Besides, statistical pattern-based mining methods cannot fully utilize the literal meaning of conceptual terms in KGs. Combining PaTeCon with human experts to efficiently and accurately obtain the constraints available for real-world scenarios is worth exploring.

\section*{Acknowledgements}
This work was supported by the  the National Natural Science Foundation of China (NSFC) under Grant No. 62072224. The authors would like to thank all the participants of this work and anonymous reviewers.

\bibliography{aaai23}

\end{document}